\def\year{2022}\relax
\documentclass[letterpaper]{article} 

\usepackage{aaai22}  
\usepackage{times}  
\usepackage{helvet}  
\usepackage{courier}  
\usepackage[hyphens]{url}  
\usepackage{graphicx} 
\usepackage{natbib}  
\usepackage{caption} 
\DeclareCaptionStyle{ruled}{labelfont=normalfont,labelsep=colon,strut=off} 
\usepackage[utf8]{inputenc}

\usepackage{url}

\usepackage{microtype}
\usepackage{subfigure}
\usepackage{booktabs} 
\usepackage{amsmath}
\usepackage{amsthm}
\usepackage{amsfonts}       
\usepackage{mathtools}
\usepackage{nicefrac}
\usepackage{algorithm,algorithmic}
\usepackage{listings}
\usepackage{balance}

\usepackage{enumerate}

\usepackage{latexsym}

\newcommand{\bfu}{\mathbf{u}}

\newcommand{\calU}{\mathcal{U}}

\newcommand{\calI}{\mathcal{I}}

\newcommand{\calC}{\mathcal{C}}

\newcommand{\tuple}[1]{\langle #1 \rangle}

\DeclareMathOperator*{\Exp}{\mathbb E}

\newcommand{\scite}{\shortcite}

\newtheoremstyle{TheoremNum}%
    {\topsep}{\topsep}%
    {\itshape}%
    {}%
    {\bfseries}%
    {.}%
    { }%
    {\thmname{#1}\thmnote{ \bfseries #3}}%
\theoremstyle{TheoremNum}

\newcommand{\commentout}[1]{}
\newcommand{\EU}{\mathit{EU}}

\usepackage{todonotes} 

\title{Modeling Recommender Ecosystems: Research Challenges at the Intersection of Mechanism Design, Reinforcement Learning and Generative Models}

\author {
    Craig Boutilier, 
    Martin Mladenov,
    Guy Tennenholtz
}
\affiliations {
     Google Research\\
     cboutilier@google.com, mmladenov@google.com, guytenn@google.com
}

\begin{document}

\thispagestyle{plain}
\pagestyle{plain}

\maketitle

\begin{abstract}
Modern recommender systems lie at the heart of complex ecosystems that couple the behavior of users, content providers, advertisers, and other actors. Despite this, the focus of the majority of recommender research---and most practical recommenders of any import---is on the \emph{local, myopic}
optimization of the recommendations made to individual users. This comes at a significant cost to the \emph{long-term utility} that recommenders could generate for its users. We argue that explicitly modeling the incentives and behaviors of all actors in the system---and the interactions among them induced by the recommender's policy---is strictly necessary if one is to maximize the value the system brings to these actors and improve overall ecosystem ``health.'' Doing so requires: optimization over long horizons using techniques such as \emph{reinforcement learning}; making inevitable tradeoffs in the utility that can be generated for different actors using the methods of \emph{social choice}; reducing information asymmetry, while accounting for incentives and strategic behavior, using the tools of \emph{mechanism design}; better modeling of both user and item-provider behaviors by incorporating notions from \emph{behavioral economics and psychology}; and exploiting recent advances in \emph{generative and foundation models} to make these mechanisms interpretable and actionable. We propose a conceptual framework that encompasses these elements, and articulate a number of research challenges that emerge at the intersection of these different disciplines.
\end{abstract}

\subsubsection*{Overview.}
This paper argues that a holistic \emph{ecosystem-oriented} perspective on recommender systems is necessary to ensure recommender systems serve the long-term interests of users, item providers (e.g., content creators, product vendors) and society as a whole. Taking economic \emph{mechanism design} as its main conceptual organizing framework, this paper outlines a collection of research challenges that we believe must to be addressed to effectively model and optimize recommender ecosystems. The paper is organized in four main parts: Part~\ref{part:intro} provides a brief introduction and a stylized formal model of \emph{recommender ecosystems} to help ground the discussion in the remainder of the paper. Part~\ref{part:whyMD} describes why mechanism design should prove to be an invaluable framework for modeling and optimizing recommender policies in such ecosystems. Part~\ref{part:nonstandardMD} outlines some of the research challenges required to bring mechanism design to bear on recommender systems, while Part~\ref{part:SCF} discusses the role of \emph{social choice functions} in making the inevitable tradeoffs in the value the recommender systems generates for different participants (e.g., users, content providers, etc.)

\part{Introduction}
\label{part:intro}

We begin with an overview of the problem, and describe a stylized model of an RS to help ground the discussion of key concepts and research problems that follow.

\section{Introduction}
\label{sec:intro}

Recommender systems (RSs) play an ever-increasing role in our daily lives, mediating the search for information, the consumption of content, the purchase of goods and services, and even the connection and communication between individuals or organizations. This ubiquity amplifies the importance of research into effective models and algorithms that ensure RSs act in the best interests of their users and society at large, and the need to integrate the resulting methods into practical RSs at scale.

The focus of the majority of RS research---and most practical recommenders of any import---is on the \emph{local, myopic} optimization of the recommendations made to \emph{individual users}. In other words, a recommendation made to one user does not consider the potential impact it might have on other users or other interested actors.
This overlooks the fact that most RSs lie at the heart of complex \emph{ecosystems}, in which the RS mediates and induces \emph{interactions} between and among users, content creators, vendors, etc. These interactions, in turn, can impact---both positively and negatively---the ability of the RS to make high-quality recommendations in the future, a fact we illustrate concretely below.

This leads to a set of questions  of substantial import pertaining to such recommender ecosystems, or \emph{recosystems}:
(1) how should we analyze and model such recosystem interactions? (2) how do we optimize RS policies in the face of such dynamic interactions? (3) what criteria and objectives should be used to guide this optimization? In this position/research challenges paper, we outline a research agenda intended to develop a deeper understanding of recosystems, and encourage research into methods, models and algorithms for the design of RS policies that maximize long-term user utility and overall social welfare across the ecosystem.

An important part of this agenda is the incorporation of concepts, methods and perspectives from economic \emph{mechanism design} \cite{hurwicz_MD:1960,hurwicz:mechanismdesign_2006,mascolell:book}. Mechanism design has played at best a minor role in the design of RSs---of course, with some notable exceptions, some of which we point to later---in large part due to the local, single-user focus typical of current research and deployment. When moving to recosystems, effective modeling and optimization of RS policies in the presence of interacting agents must account for: (i) the incentives and behaviors of the agents; (ii) the potential information asymmetry between these actors and the RS; (iii) the potential for strategic behavior; and (iv) tradeoffs in the value generated for different agents. The design of mechanisms---in our case, RS policies---in such settings is precisely the province of economic mechanism design. However, the complexity of the setting poses some unique challenges for traditional mechanism design, and we outline a number of them in this paper.

The remainder of the paper is organized as follows.
We conclude Part~\ref{part:intro} with Section~\ref{sec:omniRS} by outlining a stylized formalization of an RS---the \emph{omniscient, omnipotent, benelovent recommender}---that makes unrealistic assumptions about the knowledge and control the recommender system has regarding user preferences, content creator skill, and other factors. This model is intended to allow the crisp articulation of some of the research challenges that follow, not to reflect the full set of considerations that go into modeling RSs in the literature. Over the remainder of the article, we describe the algorithmic and modeling challenges that emerge as we relax these assumptions.

Part~\ref{part:whyMD} focuses on how mechanism design can be used to frame the design of RS policies in complex recosystems. Using the stylized model introduced in Section~\ref{sec:omniRS}, we begin in Section~\ref{sec:ecoexample} with a concrete illustration of how multiagent interactions can negatively impact the value generated by a myopic, local RS, and offer a discussion of some the considerations that can help overcome this impact. In Section~\ref{sec:MDintro}, we introduce the classical formalization of mechanism design and outline its components can help address the issues that emerge as we move from modeling recommender \emph{systems} to recommender \emph{ecosystems}.

Part~\ref{part:nonstandardMD} describes a number of research challenges associated with using mechanism design for recosystems, especially as we relax the assumptions underlying the stylized model. We outline a number of areas we think are ripe for new thinking, and make a few tentative suggestions along the way regarding specific approaches that might prove fruitful. Our research challenges touch on multiple areas of research in AI, ML, CS-Econ, and behavioral science, including: reinforcement learning, bandits and exploration, algorithmic mechanism design, incentive-compatible ML, 
user modeling, human-value alignment, generative models, ML fairness, privacy, among others.

Part~\ref{part:nonstandardMD} is broken down along the following lines: user preference modeling, elicitation and exploration is discussed in Section~\ref{sec:userprefs}. We outline the incentives and behaviors of content provider (e.g., creators, distributors, vendors) in Section~\ref{sec:creators}. 
Section~\ref{sec:info_asym} details the inherent asymmetry present in typical recosystems and how it can impact long-run ecosystem health, and in Section~\ref{sec:strategic} we address the possibility of strategic behavior, focusing on that of content providers. We conclude this part of the paper in Section~\ref{sec:dynamicMD} by describing the challenges of optimization over extended horizons.

Part~\ref{part:SCF} focuses on the inevitable tradeoffs that an RS must make when making decisions that influence the utility of multiple actors in the recosystem. The \emph{social choice} (or \emph{social welfare}) function is the standard manner in which objectives are specified in mechanism design, and it plays a crucial role here. In Section~\ref{sec:utilitySCF}, we examine tradeoffs associated with the ``direct'' utility generated for users and content/item providers by the RS, while in Section~\ref{sec:socialSCF} we discuss how the social choice function can be used to encode broader ``societal values'' (e.g., minimization of filter bubbles or polarization) and how such social dynamics might be accounted for in RS policies. We offer some final remarks in Part~\ref{part:conclude}. 

We note that this is a research challenges/position paper, and is not intended to serve as a comprehensive survey of research in this area. However, we do point to representative work that begins to address some of these challenges at various points in the paper. For a general overview of the multiagent perspective in RSs, see \citet{abdollahpouri_multistakeholder:umuai20}. There are a number of research themes in the literature that address specific aspects of multiagent modeling and optimization in RSs. Some of the more relevant for our purposes include (with representative samples of research cited):
\begin{itemize}
\item work on long-term optimization and game-theoretic analysis that models (usually relatively simple) multiagent interactions
\cite{benporat_etal:nips18,benporat_etal:aaai19,mladenov_etal:icml20,zhan_etal:www21,hron_creator_incentives:arxiv22,jagadeesan_supply:arxiv22,kurland2022competitive,benporat:aaai22,cen_trust:icmlWS22,liu_jordan_competingbandits:aistats20};
\item work on fairness in RSs
\cite{abdollahpouri_pop_bias_arxiv19,akpinar:aies22,asudeh:sigmod19,barocas_hardt:fairMLbook_2023,basu_two_sided_fairness:arxiv20,biega:sigir18,castera:acmec22,celis_fair:arxiv17,heuss:sigir22,diaz:sigir22,ekstrand_fair_recsys:handbook22,mehrotra:cikm18};
\item and work on phenomena such as filter bubbles, polarization, and popularity bias
\cite{pariser_filter_bubble_book:2011,parkes_kleinberg_opinion_dyn:kdd18,amelkin2017polar,degroot1974reaching,Ribeiro_etal:FAT20,celma:longtail2010}.
\end{itemize}

\section{The Omniscient, Omnipotent, Benevolent Recommender}
\label{sec:omniRS}

We set the stage with an intentionally stylized model of an RS that abstracts away a number of subtleties. This will allow us to illustrate a few key challenges in modeling recosystems while avoiding unnecessary complexity. It is important to keep in mind that adopting a \emph{less} stylized model---one that incorporates additional details, real-world constraints and the many nuances of agent behaviors and incentives---only \emph{exacerbates} the problems we describe, making them even more challenging. At various points in this paper, we address relaxations of this stylized abstraction and the research problems so-induced.

\subsubsection*{The Stylized Recommender.}
We assume a stylized RS, which \emph{embeds} its users $\calU$ and items $\calI$ in some latent embedding space---we equate a user $u$ and item $i$ with their respective embeddings. The RS uses cosine similarity, dot products, or (inverse) distances to measure user $u$'s \emph{affinity} for item $i$. For now, we equate affinity with user utility.\footnote{This stylized model captures, for example, many forms of collaborative filtering, including matrix factorization \cite{salakhutdinov-mnih:nips07} and dual encoders/two-tower models \cite{yiEtAl:recsys19,yangEtAl:www20}.}

For expository purposes, we assume items embody some form of \emph{content} (e.g., news, music, video) and that each item $i$ is generated by some content provider or creator $c(i) \in \calC$ \citep{grouplens:cacm97,jacobson2016music,covington:recsys16}, though everything that follows applies to other forms of item recommendation, such as product recommendation.\footnote{For instance, if the items are products for sale, one should read ``vendor'' for ``creator,'' ``sales'' for ``engagement,'' etc. The lessons below apply \emph{mutatis mutandis} to all such settings.} In general, any $c\in\calC$ can produce multiple items, but for ease of exposition, we'll assume each $c$ creates a single item (which can be altered or replaced as discussed below). Hence, we equate creators with their items for the time being. Similarly, user affinity for items typically varies with their query/request, their context, etc.\ and embeddings are often dependent on these factors; but again for simplicity, we'll treat each $u$ as single point in latent space rather than as some collection or distribution.

\subsubsection*{Omniscience.}
Realistically, these embeddings are characterized by considerable uncertainty, nonstationarity and generalization error. But we start by assuming that the RS is \emph{omninscient}. Specifically, its user and content embeddings are \emph{perfect} in the following senses:
\begin{itemize}
    \item There is no (resolvable) uncertainty:\footnote{By ``resolvable'' we refer to \emph{epistemic} uncertainty that can be eliminated with a sufficient amount of accessible data (e.g., asking the user for their preferences, or interacting with a user to gather enough data about those preferences). Irresolvable uncertainty is generally \emph{aleatoric} w.r.t.\ the model granularity and data access available to the RS.} the RS's estimated affinity for $\tuple{u,c}$ accurately reflects $u$'s (expected) utility for $c$. (This would hold approximately if, say, the models are built from sufficient amounts of data regarding both $u$ and $c$.)
    \item The model is stationary: user preferences over item space do not change (change in the content corpus is discussed below). 
    \item The model generalizes out-of-distribution: even though RS models are usually trained on the existing item corpus $\calI$, we assume that the model generalizes to new items that may be embedded at points in latent space different than any existing $i\in\calI$.\footnote{We will not focus on new users here.}
\end{itemize}
If the goal is to maximize total user utility w.r.t.\ some user distribution, under these assumptions the optimal RS policy is to \emph{match} each user $u$ to the creator/item $c^\ast(u) =\arg\max_{c\in\calC} u\cdot c$ with which they have greatest affinity whenever they engage with the system (e.g., issue a query or request a recommendation). Note that expected user utility under this policy is $\Exp_{\calU} u\cdot c^\ast(u)$, since we explicitly equate utility with item affinity. Importantly for practical RSs, the optimal matching is \emph{fully local}: the recommendation decision for one user is independent of that for another.

Naturally, since none of these assumptions hold in practice, (non-stylized) RSs will need to take steps to: (i) acquire some of this information (e.g., explore or query users about their preferences); (ii) predict this information some of this information (e.g., learn models that generalize across users and items); and (iii) optimize its matching of users to content under conditions of uncertainty.

\subsubsection*{Omnipotence.}
Assuming the user population $\calU$ is fixed (or that new users are drawn from the same distribution), we can treat the collection of user embeddings as \emph{perfect market research} w.r.t.\ content space. Specifically, one can potentially use this information to identify \emph{new content} from which users will derive significant utility.

We can draw a simple analogy with how new products are designed (e.g., see the literature on product design or product line optimization \cite{talluri_vanryzin:ms04,schon:mgmtsci2010}).
Potential consumers are surveyed about their preferences over properties or features of some feasible space of product designs. These preferences---together with constraints on the feasibility of products exhibiting specific combinations of features, and information on production (and other) costs---are used to determine the optimal product, product line, or product differentiation best-suited to a specific audience. The product is then designed, produced, marketed and sold.

Of course, our stylized RS does not actually \emph{create} content, it merely matches users to the content generated by creators.\footnote{We will discuss below the use of generative models  to support content creation (e.g., large-language models \cite{devlin2018bert,radford2018improving,thoppilan2022lamda}, image generation models \cite{vinyals2015show,esser2021taming,ramesh2021zero}) . It is not far-fetched to imagine the RS generating content itself in some cases.} Thus, the utility it generates for users is fully reliant on the \emph{content-generation decisions of creators}. However, we might imagine an \emph{omnipotent} RS that can readily induce \emph{any creator} to produce \emph{any content} the RS proposes. We might visualize this as the RS having the ability to move any creator $c(i)$ from their existing point $i$ in latent space and move them to some new point $i'$. This magically makes new content available for recommendation to users (for simplicity, we'll assume $i'$ \emph{replaces} $i$, ignoring the persistence of digital content or other digital goods).

Under what circumstances would the RS do this? There are several such conditions under which moving creators in content space can have a positive effect---for example, if the distribution of users is such that this new point $i'$ generates greater user affinity for some subset of users than $c(i)$'s current content $i$ or that of any other creator; and if the total loss in expected affinity due to $c$ moving away from $i$ is less than that gained by moving to $i'$.

Determining the placement of creators must be done \emph{jointly} of course, since the net gain associated with the move of one creator depends on the positioning of others in embedding space. Let $G$ be the space of \emph{(creator) configurations}, where a configuration $g\in G$ is a vector of embedding points $(g_c)_{c\in\calC}$ (somewhat abusing notation and treating $c$ as the creator's ``identity'' since its location $g_c$ is variable). Given a fixed configuration $g$, a user $u$ is again matched to the creator $c^\ast(u;g) = \arg\max_{c\in\calC} u\cdot g_c$ with whom they have greatest affinity. This generates total expected user utility $\EU(g) = \Exp_{\calU} u\cdot g_{c^\ast(u;g)}$. The omnipotent RS will then shuffle creators to obtain an \emph{optimal configuration} $g^\ast = \arg\max_{g\in G} \EU(g)$.

Notice that this problem can be viewed as a simple form of uncapacitated $k$-facility location (with no opening costs, hence $k$-medians \citep[see, e.g.,][]{arora1998approximation}), where each creator is a facility to be opened at a location that (collectively) best serves (or minimizes the distances across) the user population. Unlike the case above (omniscient, not omnipotent), optimization is now \emph{non-local} since it requires making utility tradeoffs across users---this is a case where genuine \emph{ecosystem interactions and tradeoffs} come into play. Few, if any, large-scale RS platforms engage in this form a global, ecosystem-level optimization.

\begin{figure*}
    \centering
    \includegraphics[width=14cm]{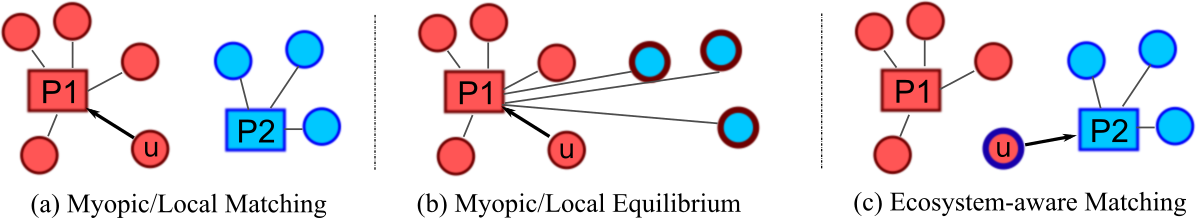}
    \caption{An illustration of (a) initial myopic (or local) RS matching of users to content/creators; and (b) equilibrium induced by myopic matching; (c) an alternative non-myopic, ecosystem-aware matching (in equilibrium).}
    \label{fig:ecosystem}
\end{figure*}

Of course, no RS can simply pick up creators and place them in ``content space'' wherever it sees fit. Creators are independent actors that have both innate and learned (or dynamic) abilities, skills, interests; specific incentives and costs for producing certain types of content; knowledge, beliefs and risk attitudes; and particular strategic, semi-strategic, myopic, or arbitrary decision-making processes. All of these factor into their selection of content to produce. As a consequence, it is vital to design RS policies that appropriately encourage or incentivize creators to make moves in content space, given their own self-interests, to improve overall user utility.

\subsubsection*{Benevolence.}
In the stylized model above, we focused on the objective of maximizing a specific form of \emph{user social welfare}, namely, the expected total user affinity for the content to which they are matched. We might be tempted to view such a RS as \emph{benelovent}, in the sense that it is doing its best to generate utility for users, including (in the case of the omnipotent RS) moving creators around to best serve users.

As appealing as this objective might be, it is untenable for many reasons, of which we list just a small selection:
\begin{itemize}
    \item Even assuming that matched-item affinity reflects user utility, simply maximizing expected affinity is inherently \emph{utilitarian} (in the Rawlsian sense), and does not capture more general notions of social welfare reflecting, say, fairness \cite{ekstrand_fair_recsys:handbook22,li_etal:recs_fairness}. This problem is even more severe under certain types of dynamics, as we show in the next section.
    \item It completely ignores the costs, benefits and incentives of content creators, who themselves derive utility from engaging with the RS, a utility which plays no role in the RS behavior described above. As we see in the next section, this can have significant, negative impacts on the long-term health of the ecosystem, including the ability to generate user utility.
    \item Utility, neither for users nor creators, is derived from a single matching, or the consumption of a single content item.\footnote{Somewhat glibly, the model above would suggest that a music RS would maximize a user's utility by playing their favorite song on repeat forever.} Instead, it must reflect the ongoing, long-term engagement with the RS. For users, this includes a number of factors, and their dynamics, such as: contexts, preferences, histories, beliefs and discovery, etc. For creators, factors include skills, beliefs, costs, incentives, etc.
    \item As discussed above, the assumption of omniscience is untenable. The RS's models are characterized by incomplete information---it can, at best, merely \emph{estimate} the preferences of users, the incentives and policies of content providers, etc. Moreover, an effective RS must generally take \emph{active} steps to reduce this uncertainty (e.g., through exploration or explicit preference elicitation) to improve the quality of the recommendations it makes.
    \item The converse of incomplete information on the part of the RS is the fact that both users and providers also lack information about the agents with whom they interact. For instance, the RS may have much greater information about the preferences of the entire population of RS users than any provider might have, since any given provider engages with only a small fraction of the population. This \emph{information asymmetry} can limit the ability of providers to make effective decisions about the content they make available, potential decreasing the value the RS can generate for users and providers alike.
    \item Finally, the behaviors of users and providers will generally be self-interested and---in the case of providers moreso than users---\emph{strategic}. An RS that fails to account for the incentives of its participants, and how these influence their behaviors, is unlikely to optimize its objectives effectively.
\end{itemize}
These factors compel us to develop a much richer modeling and optimization toolkit to manage recosystems and the interactions among its constituent actors.

\part{Why Mechanism Design?}
\label{part:whyMD}

We now turn to the the role of \emph{mechanism design} in the construction of ecosystem-aware RSs,
and specifically, why it is well-suited (at least conceptually) to address some of the characteristics identified above that challenge traditional approaches to RS design.
We begin in Section~\ref{sec:ecoexample} with an illustration of how ecosystem interactions complicate the design of RSs, using the stylized model above. In Section~\ref{sec:MDintro} we describe why mechanism design provides a valuable framework for modeling---and optimizing the design of---RSs in such recosystems.

\section{Ecosystem Interactions: An Illustration}
\label{sec:ecoexample}

Before elaborating on the role of mechanism design, we present a straightforward example to illustrate how even simple dynamics can induce complex multiagent interactions that require non-trivial optimization by the RS.

We consider a very simple, stylized scenario, based on an example of \citet{mladenov_etal:icml20}, to illustrate how ignoring ecosystem interactions can limit the ability of an RS to maximize user utility over the long run. 
In Fig.~\ref{fig:ecosystem}(a), we use (inverse) distance to reflect affinity between users (circles) and the content of specific creators (rectangles). We have two creators, P1 (red) and P2 (blue), and eight users, five of whom (red) are closer to P1, and three (blue) closer to P2.

As users request recommendations, an \emph{omniscient}, but otherwise typical, RS matches each request to the creator with which the user has greatest affinity (see Fig.~\ref{fig:ecosystem}(a)). This matching is \emph{myopic} in the sense that it maximizes immediate affinity for the user, and \emph{local} in that it does not consider impact that serving one user request might have on other users. If each user issues a single request during a specific time period, P1 engages with five users, and P2 three. However, suppose that, if a creator that does not attract four users per period, it abandons the RS---perhaps due to monetary, social or other incentives. In this case, P2 leaves after the first period, leaving the RS no choice but to recommend P1 to the three ``blue'' users at all subsequent periods (Fig.~\ref{fig:ecosystem}(b)). At this point, the system is in \emph{equilibrium}.

Unfortunately, this equilibrium leaves all blue users much worse off than had the blue creator stayed active. This clearly demonstrates that ignoring \emph{creator behavior} induced by the RS policy can have a deleterious impact on \emph{users}. By contrast, an alternative matching can be devised by anticipating the distribution of user requests--- and recognizing the (incentive-induced) behavior of the creators---that matches requests to optimize overall user welfare in the \emph{long run}. This matching (Fig.~\ref{fig:ecosystem}(c)) matches the red user $u$ to P2 and is non-local and non-myopic (it considers the impact of any user match on the long-term utility of \emph{all} users). Since $u$ is nearly indifferent to P1 and P2, it incurs a relatively small sacrifice in utility that enables a large increase in the long-run utility of the three blue users.\footnote{See \citet{mladenov_etal:icml20} for algorithms, analysis, and a deeper discussion of tradeoffs under realistic utility functions.}

This equilibrium maximizes \emph{user (utilitarian) social welfare}, and is ``optimal'' under this very specific criterion. It is also arguably more fair from an ``egalitarian'' perspective. Note, however, that it does impose a small cost on user $u$ and on creator $P1$.\footnote{See \citet{mladenov_etal:icml20} for a discussion of ways in which to distribute that cost (e.g., using stochastic matchings) or minimize user regret relative to the myopic/local matching.} Indeed, \emph{optimality} of the equilibrium (or any other outcome) depends in the choice of objective function---in the terms of economic mechanism design, the \emph{social choice/welfare function} under consideration. This argues for applying mechanism design to the construction of recommender ecosystems.

Of course, not only is this example intentionally simplified for expository purposes, it is unrealistic in the many assumptions required to induce equilibrium behavior. Among them: 
extremely simple creator preferences and dynamics; stationary, simplistic user preferences for content; no user dynamics; full information on the part of the RS about user preferences and  creator incentives/behavior; lack of strategic behavior  by creators or users;
no outside options (e.g., other RS platforms); and a simplistic objective function. Developing RS policies that work well when these assumptions are relaxed presents numerous research challenges, and is critical if RSs are to act in the best interests of users, creators, and society as a whole.

\section{Mechanism Design to the Rescue}
\label{sec:MDintro}

The example above illustrates that the design of an RS in the ecosystem context requires some form of optimization that accounts for: (i) the preferences and incentives of \emph{all} actors (e.g., users and creators) that engage with the RS; (ii) how these preferences---and the RS policy---influence their behavior; and (iii) suitable tradeoffs over the preferences of different actors. Moreover, when we relax the strong informational assumptions embodied in the stylized model above, the RS must: (i) handle incomplete and noisy information about the preferences and other private information held by these actors (e.g., their beliefs, abilities, behavioral and decision making processes); (ii) take steps to extract or refine this information when possible; and (iii) incentivize the actors to behave in a way that advances the overall objectives of the RS.

This form of optimization is precisely the province of \emph{market and mechanism design} \cite{hurwicz_MD:1960,hurwicz:mechanismdesign_2006,mascolell:book}
which essentially involves the design of policies for interacting with---and making decisions that impact---a set of self-interested agents---each of whom holds some private information---to optimize some objective that depends on the preferences of those agents.
We outline some of the ways in which a market/mechanism design perspective can help improve RS performance and play a central role in the design of RSs that lie at the center of complex ecosystems. Along the way, we suggest ways in which ``classical'' techniques need to be modified, and the required non-trivial research advances. We elaborate on a number of these research directions in the sections that follow.

\subsubsection*{Classical Mechanism Design.}
To set the stage, we provide a very (!) informal description of classical mechanism design, outlining its key conceptual ingredients.\footnote{For an overview of mechanism design, please see \cite{mascolell:book,nisan:introMD}.} We then describe how these map to entities and concepts in our recosystem setting. Concrete examples of these mappings will be provided in the subsequent sections.

Classical mechanism design assumes some action space $A$, where each $a\in A$ induces a distribution over some outcome space $O$. Each agent (or player) $g\in G$ has preferences over outcomes in the form of a utility function $u_g$, which is generally unknown to the mechanism $M$.\footnote{More generally, each agent has private information encoded in its \emph{type}, which may reflect any information not known to the designer or other agents (e.g., we might think of a content creator's ``skill'' as being unknown the the RS). However, since the role of private information is to determine the utility of an outcome to the agent, the type can be viewed as simply encoding the agent's (expected) utility function over outcomes.} The mechanism also embodies a \emph{social choice function (SCF)} $C$ that, given any vector $\overline{u} = \{u_g\}_{g\in G}$ of agent utility functions, determines the ``social utility'' $C(\overline{u},o)$ of any $o\in O$.

The goal of mechanism $M$ is to \emph{implement} the SCF by taking the action $a^\ast_\bfu = \arg\max_{a\in A} \Exp_{O} C(\bfu,o)$.\footnote{The SCF in this form is sometimes called a \emph{social welfare function (SWF)}, which induces choices, hence the SCF, via the argmax.} To do so, $M$ must take steps to observe, elicit, estimate or otherwise (possibly implicitly) determine $\bfu$ by interacting with the agents. A mechanism $M$ does so by (a) making available a set of \emph{strategies} or ``agent actions'' to each agent; and (b) specifying a mapping from the strategies/actions selected  by each agent to a choice of outcome $o\in O$. It is this strategy selection by the agents that may (indirectly) reveal something about their preferences. However, any agent $g$ may act strategically in an attempt to manipulate the selected outcome $o$ to maximize its own utility $u_g(o)$, often at the expense of $C$ (and the utility of other agents). As such, one attempts to design $M$ so that, if \emph{agent strategies are selected to be in (some form of) equilibrium}, it will (via $M$'s outcome mapping) induce the outcome dictated by the target SCF. This is known as \emph{implementation of the SCF} w.r.t.\ the equilibrium notion in question.\footnote{Common forms of implementation correspond to dominant strategy, ex-post and Bayes-Nash equilibria.}

The form such mechanisms take are many and varied, and we cannot hope to cover them here. It is important to note that not all SCFs are implementable. But one of the foundational results of mechanism design is the \emph{revelation principle}, which states (informally) that if an SCF is implementable, it can be implemented \emph{truthfully} by a mechanism which asks all participants to reveal their types (i.e., their preferences) to the mechanism. Such \emph{direct revelation mechanisms} are appealing because they restrict the actions that are made available to the agents, thus simplifying the design/policy space, and as such play an outsized role in classic mechanism design. As we detail below, however, direct revelation is completely impractical in typical recommender settings.

We now map our recosystem setting into the elements of a ``standard'' mechanism design formalization.

\subsubsection*{Agents.}
The agents (or actors) are those parties impacted by the actions of the RS, and who derive value from interacting with the RS and (possibly indirectly) with each other. Actors who have an ``interest'' in the actions of the RS (or the induced outcomes), without directly interacting with it, may be also be considered as participants in the ecosystem. Here we focus on users and content creators, but other actors of interest include content distributors, product vendors, advertisers, external organizations, regulatory bodies, other RS platforms, etc.

\subsubsection*{Actions, Outcomes and Agent Utility.}
In the recosystem, we use the term \emph{RS or system actions} to refer to those available to be taken by the RS.
In the simplest settings above, these are merely the possible one-off recommendations to users. In such cases, the RS action (in the mechanism design sense) should be viewed as the \emph{joint} set of recommendations made to all users, though we often refer to the ``individual'' recommendations as actions as well. We use the term \emph{RS policy} when emphasizing the multiple-recommendation nature of the RS's choices (e.g., across users or across time). 

We distinguish RS actions from the \emph{agent actions}, or ``strategies'' in the mechanism design sense, and often use the term ``(user or creator) behavior'' to refer to the latter.

The outcome space reflects the range of possible impacts that the RS actions can have on users and creators---again, the formal outcome in the mechanism design sense is the \emph{joint} outcome over all actors, but we abuse the term to talk about the outcome for specific actors (i.e., those aspects of the outcome relevant to that individual). In one-off recommendation, this might simply be: for a user, the actual content consumed; and for a creator, the total user engagement generated by the RS for their content.

Finally, the agent utility functions reflect their preferences over the possible outcomes---generally, this assesses only parts of the outcome ``relevant'' to them---e.g., a user's utility may depend only on the item recommended to or consumed by them, not those recommended to other users; and a creator's utility might only depend on the engagement generated by the RS for their content, and not on the engagement of other creators.\footnote{In other words, we will often assume agent utilities embody no externalities.}

The full complexity of realistic RSs naturally demands that we enrich what we consider to be outcomes induced by the RS, hence the nature of agent utility, and even the action space of the RS itself. For instance, the utility-bearing outcomes for a user of a content RS will often comprise an extended \emph{consumption stream}, or sequence of items consumed. User utility for the sequence
may not be decomposable (say, additively) into scores/utilities of the items in isolation (e.g., consider a user who values some topic diversity in the content they consume). This in turn renders the RS action/decision space much more complex:
 the RS will generally adapt its recommendations to the user's feedback or responses---hence the RS action space comprises the set of \emph{policies} that map the current history of user interactions---and in general, the history of \emph{all} users---to the choice of next recommendation to a specific user.\footnote{Even one-off recommender settings can require sequential interaction if (say) preference elicitation, example critiquing or other forms of user feedback on proposed items is allowed.}
 
For creators, outcomes may involve not just cumulative user engagement, but also temporal or other properties of that engagement over time. For instance, a creator may value some smoothness in the user traffic they serve over long periods, or user diversity across their audience.

In other domains, the (utility-bearing) outcomes of the recommendations may not be directly or immediately observable to the RS. For example, if a user accepts a product recommendation by purchasing the product, their ultimate utility will depend on the degree to which the product, its features and its functionality serves their needs over time.

\subsubsection*{The Social Choice Function.}
A mechanism's SCF embodies the relevant tradeoffs between the interests of different agents. Our stylized example above makes clear that in any RS: (i) the interests of users and creators may not fully coincide; (ii) the interests of different creators often conflict, since they are competing for audience/market; and (iii) user interests also conflict, though often indirectly via creator behavior. The specification of an explicit SCF gives the RS a clear objective when determining how utility for one user or creator should be balanced against that of another. For example: how much user attention should be directed to one creator (at the expense of others); how user utility for one group should be balanced against that of another in the near-term or long-term; and so on.

 Population-level effects may play a role in the SCF. For instance, fairness across different users or groups is usually critical, and cannot\footnote{Perhaps we should say ``should not.''  A wide swath of AI/ML fairness research does not explicitly refer to the outcomes of \emph{actions} or their utility, but rather focuses on properties of models, e.g., as in equalized odds \cite{barocas_hardt:fairMLbook_2023}.} be defined without comparison of the outcomes or utility generated for different groups. Even social values can (and perhaps should) be encoded in an SCF; e.g., the emergence of filter bubbles or polarization or other social dynamics may be penalized by the SCF. Note that this cannot be done without defining population-level outcomes.\footnote{Moreover, social value judgements are necessary if population-level outcomes are embedded in the SCF: a ``filter bubble'' in which one group of users only consumes content for ``hobby 1'' and another group consumes ``hobby 2'' may not be viewed a problematic; but if the content difference is about a particular perspective on some consequential news event, this may be treated as troublesome (e.g., because of the externalities w.r.t.\ civic discourse.)}

The use of SCFs is valuable because it forces RS designers to examine, explore, and (at least, partially) articulate the specific tradeoffs and population-level values the RS embodies and attempts to target. In complex ecosystems, RSs are unlikely to actually implement the target SCF (see discussion below of estimation, learning and optimization challenges), but the specification of core principles will undoubtedly be of tremendous value. That said, SCFs should be viewed as a ``framework'' for exploration and decision making w.r.t.\ value tradeoffs. They do not provide any guidance on what the right tradeoffs are---indeed, they cannot in general (see the impossibility results of Arrow \scite{arrow50}, Gibbard-Satterthwaite \scite{gibbard:1973,satterthwaite:1975}, and Roberts \scite{roberts:implementation1979} among others).

\subsubsection*{Preference Revelation.}
A critical aspect of mechanism design is \emph{preference revelation}---indeed, one of the primary aims of any mechanism is to induce agents to reveal private information that enables the implementation of the SCF to the greatest extent possible.\footnote{As noted above, all private information can be encoded in the utility function.} In our setting, this manifests as the RS taking actions to elicit information from---or induce behavior by---users or creators that reveal something about their preferences.

While a substantial fraction of mechanism design research concentrates on \emph{direct revelation}---mechanisms that attempt to have agents \emph{explicitly} and \emph{completely} state their preferences---such approaches are not generally feasible is RSs. Cognitive biases typically mean that users cannot reliably state their preferences accurately even in simple settings \cite{tver-kahn:1974,KahnemannTversky_CVF:2000,camerer:2003}. Moreover, the vast number of items in RSs precludes anything approaching complete revelation. When coupled with outcomes that may be dictated by combinatorial bundles or consumption streams, and the fact that user utility may be derived from some \emph{indirect} effect of item consumption,\footnote{E.g., the utility of a content item for one user may be stochastically connected to its educational impact, while for another it may increase social connection. See Keeney's \shortcite{keeney:VFT92} discussion of means vs.\ ends objectives.} we recognize that RSs will generally rely on \emph{indirect and incomplete feedback} from users. Finally, communication costs (e.g., cognitive load, potential---actual or perceived---privacy concerns) and opportunity costs (e.g., consumption of less-than-ideal items in the course of RS exploration) must be traded off against the value of any information elicited from users or creators.

Some of these factors have been studied in economic mechanism design, e.g., dynamic mechanism design \cite{parkes:onlineMDsurvey,bergemann:JEL2019}, partial preference elicitation \cite{parkes:acmec99,BlumJSZ04,hyafil-boutilier:ijcai07,lu_boutilier:aij20}, noisy responses and machine-learned preferences \cite{lahaie-parkes:acmec2004,brero_sueken_etal:acmec21}, etc., but largely in rather stylized models. Preference revelation in RSs will look quite different, relying on a combination of partial, incremental elicitation \cite{boutilier:regretSurvey2013}, attribute-based preference models to support natural (even conversational) user interaction \cite{burke-critiquing,chen_critiquing_survey:umuai2012,christakopoulou:kdd16}, generalization of preference estimation across users given observed behaviors (e.g., as in CF) \cite{Hu_etal:icdm08,koren09matrix}, bandit-style exploration \cite{auer02finitetime,christakopoulou:icdm2018,yue_etal_dueling:jcss21}, and much more.


A final important factor is that of \emph{preference construction} \cite{lichtenstein_Slovic:preferenceConstrBook2006,warren2011values}. Classic mechanism design generally assumes that preferences (and other private information) completely determine an agent's utility and are unchanging. However, users often do not have sufficient familiarity with the item corpus---nor with novel items that may not yet exist---to have fully formed preferences. From a behavioral economics perspective, we might view this as preference construction, where a user's interaction with the RS helps shape their preferences. Moreover, we should recognize that user preferences may change over time in ways that are influenced by the RS.
Naturally, similar remarks apply to content creators (e.g., their skills, interests or decision-making processes may change), item vendors (e.g., costs and opportunities may change), and other actors.\footnote{Some of what is colloquially interpreted as ``preference construction'' may be recast as changing the state of knowledge (e.g., item familiarity) of the user, which allows them to ``apply'' their well-formed preferences to item consumption differently.}






\subsubsection*{Information Sharing, Privacy and Control.}
While preference revelation is critical to RS design, we must also consider the \emph{decisions} made by users and creators. These are autonomous decision makers that, in many cases, rely on the RS to provide them with information upon which they base their decisions (e.g., consumption or creation/distribution). As such, elicitation and (indirect) sharing of some aspects of the private \emph{types} of other agents may well fall within the scope of RS's policy, and will play a critical role in improving ecosystem outcomes and social welfare.\footnote{See Myerson \cite{myerson:econ1983} for a treatment of information sharing by the principal (RS in our setting).} The information sharing of traditional RSs---recommendations to users, realized engagement to creators---offers far too narrow a communication channel for this purpose.

As one example, consider a creator (or vendor) whose item does not attain the desired engagement (or sales). In a complex RS ecosystem, the possible causes are myriad: (i) there is no demand for this type of item; (ii) item quality or other attributes make it unattractive to most users who might otherwise seek this item; (iii) numerous other creators offer similar items; and so on. The first two causes reflect user preferences while the third arises due to the ``competitive landscape.'' The most suitable creator response to each of these causes of low engagement is, of course, quite different. Unfortunately, the creator will have limited visibility into these causes, and generally be hampered in their decision making as a consequence. By contrast, the RS generally has much greater insight into both the specific preferences of the entire user population and the items of other providers that might be reducing the target item's engagement. Designing information-sharing mechanisms that allow creators to make more informed decisions could dramatically increase the ecosystem value created by RSs.\footnote{In some sense, this could be viewed as the RS providing ``market research'' to creators to enable the design of better items, as in conjoint analysis for product design \cite{green-srinivasan:1990} or product line optimization \cite{kohli:MgmtSci1990}.}

Mechanisms could include the creation of marketplaces with suggestions regarding under-served parts of the market/audience, or ``nudging'' specific creators with hints about potential improvements in their item offerings. Doing the latter effectively may require the RS to estimate (or elicit), say, the ``skills'' of the creator.\footnote{This is an example of private information that we don't colloquially consider part of the creator's utility function; but of course, formally, it can be encoded as such.}

It is essential, of course, that such mechanisms preserve the privacy of sensitive data (e.g., specific user preferences, specific behaviors and private information of ``competing'' creators). The use of aggregate audience demands, differentially private item representations
\cite{rendle_etal_privateALS:icml21,fang2022differentially,ribero2022federating}
and novel mechanisms to give users and creators fine-grained control over usage of their data are key enabling technologies that should be explored.



\subsubsection*{Strategic Behavior.}
Accounting for strategic behavior is a fundamental aspect of mechanism design. Classic mechanism design research typically assumes that agents are hyper-rational in two senses: (1) they reason about and plan their own behavior \emph{optimally} under fixed environmental assumptions; and (2) they anticipate/determine equilibrium behavior in the presence of other strategic agents. Strategic behavior clearly plays a role in RS ecosystems, as hinted above, but creators and (especially) users will generally behave in a boundedly rational fashion, a topic that gets relatively sparse attention in mechanism design
\cite{zhang_bounded_rational_MD:aer17,karaliopoulos:infocom19}.

Users will no doubt fall prey to behavioral biases that influence their choices in an RS, e.g., framing or endowment effects, hyperbolic discounting, recency bias
\cite{tver-kahn:1974,camerer:2003};
and while they may exhibit some long-term planning capabilities---potentially even seeking to influence future recommendations \cite{guo_jordan_stereotyping:eaamo21}---they are unlikely to engage in strategic reasoning w.r.t.\ the behavior of other users, creators, or the RS itself. By contrast, we expect that creators will have greater incentive to reason strategically, especially in regard to ``competition'' for market share or audience in the RS. In either case, optimal RS policies (ignoring inherent non-stationarity) should generally aim to produce approximate equilibria---either in a dynamical systems sense, a game-theoretic sense, or some mix of the two \cite{tennenholtz_rethinking:cacm19,benporat_etal:nips18,mladenov_etal:icml20,akpinar:aies22}.

\subsubsection*{Joint Optimization.}
Another daunting practical challenge in mechanism design is the computational demands of outcome selection. Even if information revelation and strategic issues are managed, outcome selection often requires solving difficult combinatorial optimization problems since outcomes must embody the features that are relevant (i.e., utility-bearing) to \emph{every} agent in the system.\footnote{For example, classical auction designs like VCG are simple in the single-item case \cite{auctions:survey}, since at most one bidder receives the item, so the outcome space is small. By contrast, \emph{combinatorial auctions} present significant computational challenges due to the induced (set of) optimization problems that must be solved for mechanisms like VCG to be used \cite{CABook}.} In a recosystem, the RS must consider the impact of its policy on the outcomes experienced by all agents in the system.

Even in one-off recommendation settings, as discussed above, this can induce a complex (offline or online) matching problem---matching millions or billions of users to thousands or millions of items and their creators---when creator behavior is influenced by the RS. As suggested above, an RS that attempts to organize or influence creator content production faces a problem akin to facility location. When user outcomes are defined w.r.t.\ entire sequences of recommendations, the optimization of the RS policy from the perspective of a single user is a form of Markov decision process (or reinforcement learning problem). Moreover, most of these problems have to be solved under conditions of incomplete information, turning (say) user sequence optimization into a POMDP. Finally, an optimal RS policy will generally take steps to gather information to improve its ability to make good recommendations, which will require computation involving quantities such as information gain or expected value of information \cite{shachter}. Most of these constituent problems are themselves combinatorial in nature, with very large inputs: the number of users, the number of creators, the number of items, high-dimensional embedding representations, etc. Taken together, these problems require considerable research to develop approximate or heuristic solution techniques that scale to practical recommender ecosystems.

\part{A Nonstandard Form of Mechanism Design}
\label{part:nonstandardMD}

We now enumerate a selection of research directions that should improve our ability to design recommender systems that work effectively in complex ecosystems. Many of the research questions that follow can be posed and addressed without appeal to mechanism design; however, some aspects of these problems---and especially the connections across these problems---are placed in clearer focus when viewed through the mechanism design lens. For this reason, we use classical mechanism design as our starting point for much of the discussion. That said, these challenges make clear that the practical design of recosystems requires that one take some license with the modelling, algorithmic and analytical techniques usually adopted in mechanism design, resulting in a decidedly ``nonstandard'' form of mechanism design.

\section{User Preferences: Modeling, Elicitation and Exploration}
\label{sec:userprefs}

Considerable RS research and practice deals with the prediction of user \emph{affinities} for items, which are used to ``score'' candidate items for recommendation.
To the extent that these reflect user utility---for instance, if the sum of the affinities of items consumed by a user reflects their  utility---we can view such RSs as attempting to uncover user preferences to better optimize its action choices, in the spirit of mechanism design.\footnote{We note that this incorporates just one aspect of mechanism design, namely preference revelation, with no multiagent preference aggregation via an SCF, nor any accounting for strategic user behavior. We will use the term ``in the spirit of mechanism design'' to refer to methods that evoke one or more elements of classical mechanism design rather than the full framework.}

\subsection{Indirect, Behavior-based Preference Assessment}
\label{sec:userBehaviorPE}

A key distinction between the typical RS approach to preference assessment vs.\ classic mechanism design methods is the fact that a user's affinities are generally estimated from the past behavior (e.g., clicks, views, or other consumption behavior) of both the user in question and \emph{other users} (e.g., using methods such as collaborative filtering \cite{salakhutdinov-mnih:nips07,koren_advances_CF:recsys_handbook2011}).
In mechanism design, the problem of incentivizing agents to reveal private information truthfully (or expend effort to support collective decision making) when this information is to be \emph{aggregated} to generate a predictive model has been studied under the guise of \emph{incentive-compatible machine learning} \cite{dekel2010incentive,balcan2005mechanism}. Adapting such models to the complexity of recosystems would be fruitful, though strategic reporting on the part of users may be rare in typical RS settings.

That said, generalization error should be accounted for in any assessment of user preferences (including when tradeoffs are to be made, as we discuss below). Moreover, the assumption that a user's behavior (e.g., choice of item from a slate) is indicative of their preferences ignores the possibility of various cognitive biases and heuristics that drive that behavior, e.g., anchoring, framing or endowment effects \cite{tver-kahn:1974,camerer:2003}, position bias \cite{joachims_position_bias:toic2007}, or even phenomena like popularity bias
\cite{abdollahpouri:aies19}
that may be partly or wholly induced by the RS itself. Given the noise, biases and incompleteness of such affinity models, we may need richer optimization techniques than those usually considered in mechanism design.

\subsection{Exploration: Indirect Revelation}
\label{sec:userExplore}

Another characteristic of the classic RS approach to preference assessment described above is the fact that no explicit steps are taken to reduce uncertainty in the RS's assessment of a user's preferences---models are trained on the ``organic'' behavior of users interacting with their recommendations. This stands in contrast to mechanism design, where the mechanism takes explicit steps that induce (possibly indirect) revelation of preferences.

Of course, considerable work has addressed \emph{exploration} in RSs, where, for example, \emph{(contextual) bandit methods} are employed to present novel items to users to construct more refined models of their preferences
\cite{li10contextual,tang2014ensemble}.\footnote{We discuss exploration further in the context of items and creators in Sections~\ref{sec:creators} and~\ref{sec:utilitySCF}.} Active exploration of this form can be viewed as incremental, indirect preference revelation in the spirit of mechanism design.

Generalization across users remains an important feature of such models. This in itself raises questions of a multiagent nature, since the value of making an \emph{exploratory recommendation} to one user---one which is not predicted to be best for that user in the current context---may lie largely in the ability of her response to improve estimates of \emph{other} users' preferences. Given that exploration imposes costs on users (at least an opportunity cost, if not annoyance, delay, etc.), interesting questions emerge regarding the potential impact too much exploration may have on user satisfaction or retention, and the equitable distribution of exploratory actions over a user population. Recent work has begun to consider strategic elements that emerge in exploration \citep[e.g.,][]{liu_jordan_competingbandits:aistats20}.

\subsection{Explicit Preference Elicitation: Direct Revelation}
\label{sec:userDirectPE}

Methods for explicit \emph{preference elicitation} have been studied extensively in the RS community, as well as in management science, operations research and many other research communities \cite{white:ismaut84,hamalainen:smc01,grouplens:active2002,preference:aaai02,toubia:jmr2004,boutilier-minimax:AIJ06,pu:AIM2008}.
The aim is to ask a user one or more questions about their preferences over items, which is akin to incremental, \emph{direct} revelation in the mechanism design sense. A variety of query types can be used to assess preferences over items. Two broad classes are those that ask about items directly and those that ask about item attributes. The former class includes rating queries (``How much do you like item $i$?''),
choice or comparison queries (``Which item from set $I$ do you prefer?''), and ranking queries (``Rank items from set $I$ in order of preference.''), each of which can be framed using a variety of modalities (text, image, GUI, speech, and combinations thereof).

By contrast, \emph{attribute queries} focus a user's attention on specific properties of the items in the corpus, e.g., color, weight, price, the presence of various product features, etc. Attribute-based methods are frequently used in multiattribute item spaces, such as shopping domains \cite{burke-critiquing},\footnote{Attributes are commonly used for example critiquing \cite{chen_critiquing_survey:umuai2012} and faceted search \cite{koren_faceted:www2008,zheng_faceted_survey:2013} which are somewhat related to preference elicitation.} but are less common in content domains, despite the fact that content attributes are often used for content search, filtering and navigation (e.g., artists, genre in music recommendation; topic or news source in news recommendation) \cite{Vig_taggenome:2012}, and sometimes for \emph{onboarding} \cite{grouplens:active2002}.

\emph{Critiquing methods} \cite{burke-critiquing,viappiani:06} can also be viewed as a form of elicitation in which more control is given to the user: when an item $i$ is recommended, the user can critique the item by requesting a different item with more, less or a diffrent value of some attribute than $i$ (e.g., a less expensive restaurant, a camera with 10X zoom, more upbeat music). Handling open-ended item critiquing will become increasingly important as \emph{conversational recommenders} \cite{christakopoulou:kdd16,sun2018conversational}---where users interact with the RS using natural-language dialogue---emerge as a dominant mode of interaction with continued improvements in foundational language modeling \cite{devlin2018bert,radford2018improving,thoppilan2022lamda}. This poses several new challenges for RSs in assessing user preferences, including: understanding how open-ended utterances reflect a user's underlying preferences; and dealing with the soft, subjective nature of attribute usage \cite{radlinksi_etal_subjective:aaai22,gopfert_etal:www22}.

Whether item- or attribute-based, \emph{interpreting} user responses to direct elicitation queries---that is, reliably, albeit stochastically, relating them to a user's underlying preferences---will have to account for the types of behavioral phenomena and cognitive biases mentioned above.

The revelation principle has meant that direct revelation mechanisms are the most widely studied within the mechanism design research community. However, considerable mechanism design research has considered the role of incremental and partial revelation, especially when the outcome space is complex or combinatorial. One example is the body of work on incremental elicitation in combinatorial auctions (see \cite{sandholm-boutilier:CAelicit06} for a survey). The principles underlying such mechanisms should play a central role in the design of RSs for complex ecosystems. This work also accounts for the strategic behavior of mechanism participants, though we expect users of most RSs to largely act in a non-strategic fashion; see the discussion in Section~\ref{sec:strategic}.

Even if users are non-strategic, they \emph{must perceive value} in the engagement they have with the RS, especially in conversational settings or when providing responses to an RS's preference elicitation queries. Each interaction imposes a potential cost on the user, reflecting, say, cognitive burden, perceived intrusiveness, or time delay in receiving a recommendation. Many elicitation techniques can quantify the \emph{expected value of information (EVOI)} of a query---how much it improves recommendation quality---which can be traded off against that cost (should the latter be quantifiable) \cite{preference:aaai02}, which can be used by the RS to decide whether to elicit further or simply recommend. However, the user must also believe (e.g., through past experience or query relevance) that the information they provide is worthwhile.

Other complexities must be dealt with when understanding user preferences, for example, the fact that most preferences are inherently \emph{contextual}---they depend on a user's current context, e.g., location, activity, companions, mood, etc.---and \emph{conditional}---the utility of one part of a recommendation depends on other parts of the recommendation.\footnote{We use the term ``contextual'' to refer to dependence on things beyond the control of the RS, e.g., a user may prefer more upbeat music when exercising, and more relaxing music when studying. We use ``conditional'' to denote a dependence on aspects of the decision or recommendation the RS can influence, e.g., a user may prefer a late reservation at one restaurant, but an earlier reservation at another.} Sequential recommendations are an important special case discussed below.


\emph{Conversational recommenders} \cite{christakopoulou:kdd16,sun2018conversational} have been proposed as a means to allow more flexibility in a user's interactions with an RS, including allowing more open-ended dialogue to support richer forms of steering, critiquing, preference elicitation, and user probing/exploration, and at the same time allowing the RS to develop a more nuanced understanding of a user's preferences and context to support many of the considerations above. 
The rapid development of generative AI and foundation models \cite{devlin2018bert,radford2018improving,thoppilan2022lamda,vinyals2015show,esser2021taming,ramesh2021zero} holds promise for building highly performative conversational RSs. For example, the use of large-language models (LLMs) to support multi-turn dialogue-based recommendation \cite{friedman2023leveraging,gao2023multi} offers an immediate opportunity to allow more open-ended preference expression, critiquing, preference elicitation and recommendation explanation. At the same time, significant challenges remains, including developing models that are inherently personalized, and that seamlessly blend the rich, behavior-based models of users and items commonly used in CF-based RSs with the rich semantic understanding of items and users afforded by LLMs. Incoporating multi-modal interactions into RSs also represents rich opportunities, e.g., using text-to-image models \cite{ramesh2021zero} to synthesize new content or stylistic variations of products to allow more efficient user exploration or critiquing, or using audio models to generate new forms of preference elicitation queries.

\subsection{Unobservable Outcomes}
\label{sec:userUnobservable}

The discussion of preferences above focuses on the assessment, elicitation and revelation of preferences or utilities for individual items and makes several unrealistic assumptions. One of these is that fact that RSs usually \emph{equate the item and the outcome}. In other words, they treat the consumption, purchase, or other engagement with the item by the user as the outcome over which preferences should be assessed.

There are a variety of situations in which this is not the case. In particular, often the utility-bearing outcomes for a user are merely \emph{facilitated} by the item recommended. This embodies a distinction between what Keeney \scite{keeney:VFT92} famously called \emph{means objectives} vs.\ \emph{ends objectives}. While an RS talks about items in terms of their features, user’s often derive value from the usage of the item facilitated by those features. For example, an RS might help a user navigate the purchase of camera by discussing various features and technical specifications, but its value may ultimately depend on whether its intended use in primarily for vacation pictures, nature photos, or great action shots of kids on their sports teams. Understanding user preferences over outcomes (the ends) may be more natural than a focus on item features (the means), especially when the mapping between the two is complex and better understood by the RS.

This also implies that user utility is often influenced by exogenous factors beyond the control of the RS or the user. For example, a restaurant recommendation may be better or worse for a specific user depending on the availability of parking nearby and the nature of the user's commute. Moreover, the value of a reservation at a specific time---should the restaurant have highly variable service times---might depend on whether the user has after-dinner theatre tickets. The outcomes are stochastic, and such factors require assessing user preferences and risk attitudes in \emph{that outcome space}, as well as having an understanding of domain dynamics.

Finally, \emph{latent factors} may play a dominant factor in user outcomes. For example, a user's true satisfaction with (or enjoyment derived from) a music playlist will generally not be observable. Such latent factors may be reflective of (or even equated with) user utility, and some practical RSs have started incorporating regular user surveys to assess and predict factors like user satisfaction with recommended items \cite{goodrow_youtube_blog_2021}. Of course, it may be difficult for a user to express such factors with any degree of confidence or precision when asked. Other latent factors may be more directly outcome-related, e.g., the consumption of news content to stay current or informed on a specific issue, or entertainment content to facilitate social interactions with friends. Here the user's \emph{state of knowledge} may be latent, but changes in that state may be (part of) the outcome of interest.

Notice that each of these factors necessitates a reconception of the outcome space relative to traditional RS designs.

\subsection{Sequential, Multi-item Recommendation}
\label{sec:userSequential}

Finally, in most recommendation domains, typically users engage with a specific RS repeatedly for the recommendation of multiple items. This is especially true in many content domains (e.g., listening to music, reading news, watching video content), but in others as well (e.g., shopping, commerce, entertainment and dining).\footnote{Exceptions are largely confined to recommendations involving one-off or infrequent decisions (which tend to be higher-stakes).} In some cases, utility may derived over a single session (e.g., a music listening session), while in others it may extend over long, multi-session horizons (e.g., consumption of news or educational content, or numerous music sessions intended to broaden one's musical appreciation). As such, user value derived from the collection or sequence of recommendations is not simply the sum of individual item utilities, and will generally require some form of sequential optimization using techniques such as reinforcement learning (RL) or other nonmyopic planning methods.

In some cases, sequence utility may be a function of individual ``rewards,'' but exhibit a certain \emph{risk profile} w.r.t.\ the sum of rewards \cite{keeney-raiffa,mihatsch:mlj2002,chow_etal:jmlr2017}, diminishing returns, or even the classic sigmoidal utility form of prospect theory \cite{kahneman_prospect:1979,prashanth:icml16}. For instance, user utility or satisfaction for video content may be convex in the region of low cumulative reward (a large amount of low reward content may be viewed as unsatisfying, or a waste of time) and concave in the region of high cumulative reward (saturation or diminishing returns).

In other cases, utility may not even be a function of the isolated scores or rewards of the individual items; it may instead depend on other properties of the items in the sequence and their relationship. A classic example is content diversity, where a user may value topic or stylistic diversity in a consumption stream. In this case, the quality or reward of individual items is not sufficient to assess the utility of the sequence; the distribution of diversity-relevant attributes across the sequence must also be considered.

Preference assessment and elicitation over sequences is generally much more involved than for single (even multi-attribute) items. While techniques such as elicitation of reward functions in Markov decision processes (MDPs) \cite{Regan-Boutilier_MDPelicitation:ijcai11} and preference-based RL \cite{wirth2017survey} may offer some insights, richer techniques will be needed in complex recommender domains. Accounting for behavioral biases \citep[e.g., hyperbolic discounting][]{laibson_hyperbolic:qje97,laibson_temporal:handbookBE2019} and risk preferences \cite{charness_riskprefs:jebo13} in the assessment of long-term outcomes should prove critical as well.
Elicitation and assessment over recommendation sequences is also relevant to the discussion of dynamic mechanism design in Section~\ref{sec:dynamicMD}.

\section{Creator Incentives, Skills and Beliefs}
\label{sec:creators}

The providers of the items in the RS's recommendable corpus (e.g., content creators, distributors, item vendors) have their own incentives for making these items available, which influence their behaviors. These incentives could be economic or monetary: product vendors have direct monetary interest in recommendations being made to users to whom their products appeal, while content creators may have a direct monetary interest in user engagement (e.g., product placement or ad revenue), or indirect economic interest in such (e.g., eventual monetization of reputation and brand). Incentives might also be driven by social, influence or reputational factors, e.g., a content creator who derives utility from the broader influence of their ideas. User engagement may serve as, at best, an indirect proxy for creator utility.

\subsection{Creator Preferences and Incentives}
\label{sec:creatorPreferences}

In contrast to the detailed modeling of users (albeit still limited as discussed above), most RSs do much less detailed modeling of provider incentives and behaviors, and in many cases, none at all.\footnote{A notable exception is work in online advertising, which we discuss briefly below.} As our stylized example in Section~\ref{sec:ecoexample} illustrates, failing to account for creator behavior can have a significant impact on the utility an RS can generate for its users. Moreover, a good understanding of creator preferences allows the RS to \emph{better incentivize alternative behaviors} (e.g., provision of new products, generation of new content) that can improve overall user utility over time. Finally, given that creators are themselves actors in the system, RSs should generally balance the utility creators derive from engaging with the RS against each other and against user utility \cite{abdollahpouri_pop_bias_arxiv19,akpinar:aies22,asudeh:sigmod19,basu_two_sided_fairness:arxiv20,biega:sigir18,celis_fair:arxiv17,SinghJoachims:kdd18,heuss:sigir22,diaz:sigir22,ekstrand_fair_recsys:handbook22,mehrotra:cikm18}. We discuss each of these issues in turn.

Models of provider utility seem to be used infrequently in practical RSs and incorporated into RS research in fairly simple ways. To the extent that such models exist, they tend to be of a relatively simple form, for example, using product sales \cite{wang2015revenue,louca2019joint}, user impressions/views or the like as a proxy for utility . For example, the body of work on \emph{fairness of exposure} focuses on user impressions across different creators \cite{SinghJoachims:kdd18,heuss:sigir22}, though without explicitly labeling this as creator utility (see Section~\ref{sec:utilitySCF}). Different forms of creator utility are almost certainly at play in real recommender ecosystems. For example, a creator's utility for engagement with their content may be a complex function of total, cumulative engagement, e.g., exhibiting specific risk profiles, diminishing returns, sigmoidal structure and saturation, etc.  \cite{keeney-raiffa,kahneman_prospect:1979}. This may be especially true when factoring in item production or sourcing costs, etc. Even factors such as user demographic mix, the distribution of user engagement across their audience, and related user properties can play a role (as can indirect outcomes, as discussed in Section~\ref{sec:userUnobservable} in the case of users). Finally, properties of the temporal sequencing of engagement or conversions may play a direct role (e.g., smoothness).

Given the potential complexity of creator utility, considerable research is needed to develop both indirect preference assessment and exploration techniques, as well as explicit preference elicitation methods. Indirect methods will of course require understanding how a creator's preferences are manifest in their behavior (see Section~\ref{sec:creatorBehaviors}).

As we discuss in Section~\ref{sec:strategic}, unlike users, we expect creators to be somewhat \emph{strategic} in their decision making. This then requires designing (indirect or direct) elicitation mechanisms in the spirit of classic mechanism design: the mechanism itself should either account for the game-theoretic responses of creators or provide incentives that minimize the degree to which creators misrepresent or misreport their preferences. In the case of indirect signals (e.g., the type, quality or frequency of new content generation; pricing and type of new product offerings), connecting these to the creator's underlying preferences in equilibrium becomes critical. This may be especially challenging when monetary compensation is not allowed, not only due to classic impossibility results in the theory of mechanism design \cite{gibbard:1973,satterthwaite:1975,roberts:implementation1979}, but also because of the need to learn and generalize behavioral models across creators.

One area in which the principles of mechanism design have been applied with great success is computational advertising \cite{mehta:onlinematching2013,dave2_varma_advertising:ftir2014}. If we (somewhat loosely) view ad servers as RSs---where the ads of specific advertisers are recommended to users depending on the relative affinity a user has for an ad (or the product and services they represent)---this serves as a successful example of creator (advertiser) incentive modeling and elicitation. Not only do classic auction mechanisms or their variants play a central role (first-price, second-price, generalized VCG, etc.), methods such as budget smoothing, ROI metrics w.r.t.\ conversions, and a variety of other mechanisms can be interpreted as adding at least some nuance to the usual ``additive'' clicks/conversions view of utility. However, apart from ad ``relevance,'' little attempt is made to model user utility to any degree; and most modeling---whether pCTR or conversion predictions, or even more subtle concepts such as ad blindness \cite{hohnhold:kdd15}---tends to focus on user behavior rather than user utility. Finally, from a creator perspective, preference modeling and elicitation---including interpersonal comparisons of utility across creators---is greatly simplified by the use of quasi-linear utility, where money serves as the common numeraire. That said, the adaptation of techniques from online advertising to mechanism design for recommender ecosystems holds considerable promise.

\subsection{Creator Behaviors}
\label{sec:creatorBehaviors}

Understanding creator preferences is obviously important when these are incorporated in the SCF. However, even if the SCF inputs consist solely of user preferences, since creator behaviors are shaped by their incentives, creator preferences themselves impact the ability to optimize a ``pure user'' SCF as we showed in Section~\ref{sec:ecoexample}.

The behaviors or decisions of any rational actor are based not only on their preferences, but also on their available actions and their beliefs. Hence, modeling creator behaviors may, in many cases, be best served by modeling both:
\begin{enumerate}[(a)]
    \item their \emph{action spaces}. For example, what skills and resources to creators have to generate new content, of specific type and specific quality? What abilities do vendors have to source, design and/or manufacture new products and at what capacity and cost?
    \item their \emph{beliefs}: For example, what do creators believe (e.g., are they able to predict) about the audience for a new content item, the time and cost required to generate attractive new content for an existing or new audience, or the anticipated quality of the results? What does a vendor believe the market will be for a new product as a function of its price?
\end{enumerate}

Predictive models of creator behavior may be feasible without explicit modeling of skills and beliefs, at least in simple cases. For instance, our stylized example assumes creators abandon the RS if they do receive sufficient user engagement with their content. Note that this model of behavior does not explicitly capture any form of utility, but could be relatively simple to learn (given, say, sufficiently exploratory and generalizable data). However, the underlying explanation for abandonment w.r.t.\ creator incentives and costs, available actions/responses, and beliefs, allows for a much richer set of robust interventions by the RS with more predictable effects. For example, a creator might abandon the RS because they believe there is little audience for the type of content they produce, or because they believe other creators offer higher quality content of the same type. How the RS responds to keep the creator engaged (either to improve utility for that creator, drive value for users, or both) depends on the rationale for their abandonment decision. We discuss this further in Section~\ref{sec:info_asym}.

Developing models of creator skills and abilities, of creator beliefs, and of creator decision-making policies---especially those that generalize across creators---present a number of interesting research challenges. Among them are: developing models of the appropriate form; exploration and elicitation methods for indirect/direct assessment of these models; and appropriate incentivization for revealing this private information when creators deliberate strategically.

\section{Information Asymmetry and Ecosystem Health}
\label{sec:info_asym}

The ``health'' of a recommender ecosystem is often discussed without precise definition, but is typically evoked to refer to the ability of an RS to generate diverse recommendations to its users. We take a somewhat broader, though still informal, view, equating recosystem health with \emph{the ability of the RS to
generate significant value for all of its participants over the long run}. This may encompass a variety of factors relating to the total utility or welfare it generates for users, creators, vendors, etc.; the distribution of this utility; the facilitation of user well-being; the ability to generate beneficial social dynamics; and much more. Many of these factors may, of course, incorporate diversity as one component (see Part~\ref{part:SCF}).

Importantly, we think of health not as a ``snapshot'' of utility generation, but rather as the RS's ability to anticipate and respond to fundamental changes in the underlying ecosystem, for example, as user preferences and tastes evolve, as new content creation or product production capabilities emerge, or as user/creator communities form and disband. Moreover, we expect an RS to take actions to promote, or at least facilitate, beneficial changes.

Since the value created by the RS depends on both the production/provision decisions of item providers (e.g., content creators, vendors) and the consumption decisions of users, it can often impact value creation most directly by supporting this decision making, encouraging decisions that maintain ecosystem health and promote long-term social welfare. Here we focus primarily on content creation (or item production) decisions to illustrate some key insights and research challenges.

One of the impediments to ecosystem health is the nontrivial \emph{information asymmetry} that exists between the RS and content creators. First, we note that the RS has rich models of user preferences over the existing item corpus. If we assume that this model generalizes \emph{out-of-corpus} to some degree, we can treat this as a comprehensive model of \emph{latent user demand} over content space which predicts affinity or utility for \emph{hypothetical items} that do not (yet) exist. Second, the RS has a holistic view of the corpus itself, and potentially some insight into the abilities of providers to source or create new items (see discussion above of creator skill). This serves as a deep understanding of both the current and \emph{potential} supply of items for recommendation and consumption. Unfortunately, no provider has the same breadth of insight into global demand or supply, since they tend to interact with only a small subset of the user population. This information asymmetry limits the ability of creators to make informed content generation decisions, and is a key source of economic inefficiency.

To illustrate (see discussion on Section~\ref{sec:MDintro}), consider a creator whose content does not attain the desired user engagement: as discussed above, possible causes include: (i) no demand for this type of content; (ii) content quality makes it unattractive to most users; or (iii) numerous other creators offer similar items. While the RS has information that distinguishes these causes, the creator may not. In most RS settings, creators are left to their own devices to explore, design and test new offerings---a process that is generally inefficient, costly, and sometimes disincentivizing.

Breaking the information asymmetry through some form of direct or indirect information sharing can improve the decision-making capabilities of creators and drive significant improvements in both user and creator utility. For instance, if the competitive landscape is limiting a creator's audience (cause (iii)), the RS could either communicate this directly, or provide more indirect guidance by ``steering'' the creator to produce content in a less well-supplied part of content space for which latent demand is predicted to be high.

We can interpret the RS demand and supply modeling as a form of ``indirect'' market research which can be shared with creators to enable the production of items that better meet untapped user demand. But a variety of design and research challenges must be resolved to make this happen. We list a few of these here:
\begin{itemize}
    \item Direct information sharing may not be feasible in many domains, for example, if it reveals personal user data, strategically important information about ``competitors,'' or sensitive information about RS policies. Indirect or \emph{implicit} sharing may be more acceptable; e.g., in our example above, revealing the predicted, aggregate audience for new content in an effort to support a creator's content creation decisions. Of course, the leakage of private or sensitive information \cite{Dwork} must be safeguarded in any such mechanism.
    
    \item Sharing information alone may not be sufficient to induce welfare-improving changes in content generation by a creator. The costs and uncertainty (hence risks) associated with major changes may serve as a disincentive for a creator to generate truly novel content. For example, the creator's lack of experience, skill or access to resources, their uncertainty regarding the size/value of the new audience, and their potential lack of trust in the RS's ``advice'' might all need to be overcome. Hence, appropriate mechanisms must be investigated that: incentivize new content production; de-risk creator exploration; facilitate the development of new skills; and open new audiences---all while maintaining trust in the RS's suggestions and analytics. Many such mechanisms (e.g., gradual exploration, skill development, access to resources) may extend over lengthy horizons.
    
    \item Models of user preferences generally rely on behavioral interactions with items in the corpus, content features of these items, or a mix of both. Proposing ``hypothetical'' items to creators thus presents some significant challenges. For example, a collaborative filtering (behavioral) model will generally embed users and items in some latent space (see Figure~\ref{fig:ecosystem}). If some point in content space is predicted to have high user demand, mechanisms for generating actionable descriptions or prompts for creators is critical. There may be a significant role for generative models, such as LLMs \cite{devlin2018bert,radford2018improving,thoppilan2022lamda}, or text-to-image models \cite{vinyals2015show,esser2021taming,ramesh2021zero}, to synthesize language descriptions, item features, evocative images and the like to serve as effective guidance for the creator.
    
    \item RSs may also play an active role in content creation. There is an clear trend toward placing generative AI and other tools in the hands of creators \cite{gillick_eck_etal:icml19,yang2022diffusion}, and such approaches could be extended to support explicit product or content ``design.'' More than this, RSs are in a unique position to facilitate a full design-create-test-refine loop for creators or designers, coupling direct \emph{experimentation and analytics} into the iterative design/creation process. In addition, such experimentation could involve  preference elicitation with willing users to serve as a form of \emph{direct} market research.
    
    \item While the discussion of information sharing and advising on content generation above has emphasized individual creators, the process is in fact one that has to be \emph{coordinated} across the entire set of creators. Suppose the RS proposes generation of novel content to one creator that suggests a certain audience can be reached. A similar suggestion to a different creator might undercut the audience suggested to the first. As such, we must think of the problem of ``nudging'' creators to new parts of content space as a large-scale \emph{joint optimization} problem \cite{prasad_etal:creators2023}---see our discussion of the Omnipotent Recommender in Section~\ref{sec:omniRS}, where we loosely characterized this as a facility location problem. Of course, the problem is rendered even more complex by the recognition that prompting creators is itself a long-horizon, multi-stage process. Moreover, there are subtleties associated with the potential to promote \emph{strategic competition}: unlike facility location, where the most effective solution tends to ``spread out'' facilities, in an RS ecosystem, the ideal configuration of production may actually keep several creators ``near'' each other, since some degree of competition for user engagement may actually incentivize the production of higher quality content.
    
    \item Once an RS takes steps to ``nudge'' providers to certain points in content space, the incentive for explicit strategic behavior may, in fact, increase. For example, one creator may refuse to change its content strategy when prompted if it believes that this refusal might induce a competing creator to make such a change (see discussion of Strategic Behavior by creators in Section~\ref{sec:providerstrategic}).
    
    \item Rather than nudging specific creators---a process that requires estimating (or eliciting) a creator's incentives, skills and resources--- another class of mechanisms for (indirect or direct) information sharing could include the creation of semi-open marketplaces that communicate under-served parts of the market/audience to \emph{any} creator interested in filling that gap. This is itself an interesting market design and optimization problem that should target desirable (long-run) equilibrium behavior.
    
    \item Finally, any mechanism that engages in direct or indirect information sharing must take pains to avoid \emph{sharing} sensitive or private information about users or other creators, or \emph{leaking} such information when providing any aggregate data, metrics, prompts, suggestions, etc. 
    (for instance, in the sense of differential privacy \cite{Dwork}). 
    This includes direct information about the preferences or behaviors of specific users that a creator does not already have access to, or revealing information about a competitor's incentives, strategies, skills or cost structures. While differentially private methods have been explored in standard RS settings \cite{liu_dp_CF:recsys15,rendle_etal_privateALS:icml21,carranza_private_llm_RS:arxiv23}, extending such techniques and analyses to complex recosystems is needed.
\end{itemize}







\section{Strategic Behavior}
\label{sec:strategic}

By \emph{strategic behavior}, we adopt the standard informal economic/game-theoretic notion, referring to actions taken by an agent that anticipate the actions or reactions of other agents. Our discussion above has largely assumed that users and creators behave non-strategically. We turn to what strategic behavior might look like for both users and content creators (or other item providers). 

\subsection{User Strategic Behavior}
\label{sec:userstrategic}

We expect users, for the most part, to behave non-strategically. So when a user is presented with several content recommendations, they will simply select their most preferred option given their current context, possibly subject to random noise. In sequential settings, such a choice may be more challenging since the value of the selected item may depend on future recommendations/selections, and may influence the recommendations the user receives in the future. Accounting for the first factor requires that the user behave sequentially rationally (i.e., plan); but since the RS policy may be influenced by the initial selection, a user may also invoke some ``mental model'' of the RS policy and thus behave in a way to explicitly influence subsequent recommendations \cite{guo_jordan_stereotyping:eaamo21}.

The latter could be viewed as simply planning in the environment ``determined'' by the RS policy, and thus remains non-strategic---this would be the case if the user believes the RS policy is stationary and/or attempting to act in the user's best interests. For example, a user might select a (non-preferred) track by a specific artist in a music RS because they believe it will induce the RS to propose additional (preferred) tracks by that artist moving forward. If by contrast the user believed the RS were trying to promote specific content independent of the user's interests, they might try to take actions to prevent that, in which case, we might view such behavior as strategic.

Other forms of strategic behavior might involve spam-like activity to promote the popularity of a favorite musical artist, news outlet, or content creator; or to provide excessive ratings or glowing product reviews to increase the odds of certain items being recommended to others. Likewise, responses to preference elicitation queries may intentionally be inaccurate to manipulate the RS's future recommendations to the user in question, to other users, or to impact the providers of the recommended items (either positively or negatively). Handling potential mispresentation of preferences during explicit elicitation is the primary objective of classic mechanism design approaches, though as discussed in Section~\ref{sec:userDirectPE}, any mechanism design approach must accommodate a very complex outcome/preference space and the incremental, incomplete, noisy and biased nature of practical user preference assessment. With regard to indirect preference assessment (see Sections~\ref{sec:userBehaviorPE} and ~\ref{sec:userExplore}), the somewhat nascent body of work on \emph{incentive-compatible machine learning} \cite{dekel2010incentive,balcan2005mechanism} is certainly relevant, though currently far from meeting the practical demands of complex recosystems.

All this said, user strategic behavior, if existent at all, is likely to be relatively limited. Often it will involve trying to promote the interests of favorite item providers, and hence be generally somewhat ``cooperative.'' Moreover, we expect users to exhibit extreme bounded rationality w.r.t.\ the full complexity of the RS ecosystem---they will rarely consider the true equilibrium state of the system.

We note that the increased used of explicit \emph{two-way} communication about a user's preferences as advocated above---both in terms of users providing explicit preferences and critiques, and the RS offering explanations and exploration to users---is likely to increase transparency and user trust. This, we believe, will reduce the incentives for users to expend cognitive effort in attempts at ``non-strategic'' manipulation of RS behavior.

\subsection{Provider Strategic Behavior}
\label{sec:providerstrategic}

Strategic reasoning is more likely to play a role in the behavior choices of item providers, such as content creators or product vendors, than it is with users. Disregarding any direct intervention by the RS, a product vendor will generally choose to offer products for sale that it not only predicts to have reasonable demand, but that are sufficiently differentiated from the offerings of other vendors, including those that others may produce or source in response to the choice of the vendor in question. Price setting will likewise be strategic. Similar considerations will generally play a role in content generation by creators even if RS users do not (directly) pay for that content.

An RS that takes actions to maximize some form of social welfare must account for provider strategic behavior. The RS matching policy itself can induce strategic behavior on the part of content providers, a problem that has been analyzed by \citet{benporat_etal:nips18}. Extensions of this modeling approach to richer user and creator preference models would have tremendous import.

From the perspective of more direct elicitation of private information, consider our illustrative example where each provider requires some minimum audience engagement to continue making content available. Directly eliciting this target engagement clearly gives the provider a \emph{prima facie} incentive to overstate their target, but not by too much.\footnote{Notice that while demanding a larger audience increases the odds of the RS recommending a provider's content to more users, it also runs a greater risk of being deemed infeasible given the demands of other providers, and hence shutting out the provider from being matched at all \cite{mladenov_etal:icml20}.} Direct revelation mechanisms of this sort are akin to auctions, where the matching should account for equilibrium behavior of providers (e.g., as in a first-price auction). But in many RSs (e.g., content RSs) monetary transfer cannot be considered, so this class of problems falls squarely within the area of mechanism design without money \cite{SchummerVohra:facilityLocSurvey,Procaccia-Tennenholtz:acmec09}, which can be especially challenging to deal with.

Next consider the setting described above in which the RS
takes steps to induce providers to generate (or otherwise make available) items that will improve overall social welfare in the ecosystem. Here again, provider strategic reasoning is likely to emerge, and should be taken into account. For instance, if creators $C_1$ and $C_2$ are producing content well-suited to the same audience, they may split the audience, receiving less engagement than if their content were more distinctive. To ameliorate this, the RS might prompt $C_1$ to make slightly different content, to the benefit of the broader user population as well as both creators. However, $C_1$ might refuse to do so in the hopes that the RS might then request that $C_2$ move instead, thus sparing $C_1$ the potential cost and risk of this change.

Incentivizing welfare-improving behavior across a diverse set of strategic or semi-strategic providers will require significant effort involving many aspects of mechanism design. While direct mechanisms may work in relatively simple domains (e.g., auction-like mechanisms), in complex settings we expect that more detailed modeling of provider private information---utilities (e.g., w.r.t.\ induced engagement), skills and costs, beliefs, etc.---and how these elements shape their strategies and behaviors---especially when providers exhibit various forms of bounded rationality---should ultimately prove most effective. See also our discussion above of information sharing, which should take into account the potential strategic effect it has on these actors \cite{myerson:econ1983} (providers in our case).

\section{Dynamic Mechanism Design and Reinforcement Learning}
\label{sec:dynamicMD}

The \emph{sequential} nature of an RS's interactions with users, creators, vendors and other actors brings with it a host of challenges for the mechanism design perspective. This includes questions regarding preference elicitation (see Section~\ref{sec:userSequential}), learning and optimization. First turning to learning and optimization, constructing optimal sequential interaction policies has attracted a fair amount of attention within the recommendations literature, where a significant body of work has formulated sequential recommendations using MDPs or partially observable MDPs \cite{shani:jmlr05} or tackled the problem using reinforcement learning \cite{taghipour:recsys07,choi2018reinforcement,facebook_horizon:2018,chen_etal:2018top,slateQ:ijcai19}. However, the bulk of this work focuses on ``local'' policies that optimize for one user in isolation. Extending RL methods to handle the large-scale interactions present in recosystems will generally require joint optimization of the form studied in multiagent RL \cite{busoniu2008multi,lanctot2017unified}. The study of such \emph{multiagent sequential} optimization for recosystems is currently in its infancy \cite{mladenov_etal:arxiv20,zhan_etal:www21}.

Turning to incentive issues and information revelation---two key ingredients of mechanism design---sequential interactions are the subject of
\emph{dynamic mechanism design (DMD)} \cite{parkes:onlineMDsurvey,athey:econ2013,bergemann:Econ2010,bergemann:JEL2019,SCB-multistage:ijcai07,curry_etal:ewrl23}, which provides a framework for mechanism design where different agents can engage with the mechanism at different points in time and possibly over varying horizons. Most of the work in DMD addresses problems involving pricing in dynamic markets (e.g., dynamic pricing for tickets, surge pricing for ride-share programs, wholesale energy markets), and adapt mechanisms like VCG to such settings. 

Dynamic mechanisms of this sort may provide useful foundations for mechanism design in recosystems, but cannot be applied directly for a variety of reasons. One of these is the fact that the focus on pricing problems allows the use of monetary transfers to incentivize suitable behavior and information revelation. But, as discussed above, RSs often fall within the realm of ``mechanism design without money,'' a problem that has been studied in the challenging dynamics setting only sporadically \cite{guo_dynamicMD_without_money:2015,balseiro_multiagentMD:or2019}. A second is the fact in the dynamics of complex recosystems is rarely known, and often must be learned. Recent work has explored the use of RL to learning optimal dynamic mechanisms \cite{lyu_RL_dynamicMD:arxiv2022,lyu:icml22} from repeated interactions with the environment. While directions such as these are promising, nontrivial research is needed to realize the potential of DMD for modeling recosystems.

\part{Tradeoffs and the Social Choice Function}
\label{part:SCF}

As discussed above, the ecosystem perspective makes it abundantly clear that an RS must make tradeoffs in the utility it generates for different actors in the system: users, creators, vendors, etc. Except in systems with the most trivial outcome spaces, the incentives of these parties will rarely be fully aligned. While there are many types of incentive conflicts, and a variety of ways to ``carve them up,'' we consider two classes of tradeoffs here: those involving the preferences of the individual actors (Section~\ref{sec:utilitySCF}); and those that we refer to as ``social tradeoffs,'' pertaining to the properties of the RS policy that may be less easily articulated in terms of an SCF over the preferences of individual participants (e.g., these may be interpreted as imposing externalities beyond the RS ecosystem itself). We elaborate on this distinction below.

\section{Tradeoffs over Actor's Utilities}
\label{sec:utilitySCF}

One of the primary calling cards of mechanism design is the use of an SCF to encode the tradeoffs the designer is prepared to make between the utilities generated by the mechanism for its participants. For example, maximizing utilitarian social welfare generally requires that some agents receive greater utility than others.

Our stylized example in Sec.~\ref{sec:ecoexample} illustrates this in the case of RS ecosystems, where there is a fundamental tradeoff in the utilities that can be generated for different users---we generate a large increase in utility for the blue users at a small cost in utility for the red users under the ``user utilitarian'' optimal equilibrium policy. Similar remarks apply to the providers themselves: the ``user utilitarian'' optimal policy changes the utility mix among the providers, by more evenly distributing engagement across the red and blue providers. If provider utility is equated with engagement, we note that this too improves total provider utility (hence ``provider utilitarian'' welfare). Of course, under different utility functions, user utility and provider utility may not coincide in this fashion. 

The use of mechanism design in the design of RS policies requires the specification of an SCF. This forces the RS designer to explicitly articulate the tradeoffs they are prepared to make---which outcomes, across all actors in the system, they consider more or less desirable---rather than leaving them to chance. A variety of factors can be brought to bear in the construction of an SCF. Several broad classes of social welfare functions include:
\begin{itemize}
    \item Utilitarian SCFs, which maximize the sum of (expected) utilities generated over all agents (e.g., users, vendors), or some subclass (users only). One criticism of a pure utilitarian approach is that it can generate wide disparities in utilities across agents (e.g., users or providers), which some may deem unfair (we discuss fairness further below). For instance, in our stylized example, if none of the red users were sufficiently close to the blue provider, the utilitarian-optimal policy would not have ``subsidized'' the blue provider, thus leading to very low utility for the blue users (and no utility at all in steady state for the blue provider). 
    \item Maxmin SCFs, which maximize the utility of the worst-off individual (with least utility), either across the population or within some subclass. While seemingly more egalitarian, maxmin SCFs can sacrifice a significant amount of utility for a large number of agents to accommodate the worst-off agents. 
    \item SCFs with fairness and/or regret constraints or objectives. These can optimize some primary objective (say, sum of utilities), while constraining the solution in one or more respects. Fairness constraints tend to encourage something akin to maxmin outcomes (ensuring outcome utilities are somewhat less disparate), while regret constraints can be used to penalize policies that impose too large a cost (loss in utility) on any individual in service of the primary objective \cite{mladenov_etal:icml20}. Fairness is discussed further below, while regret measures require one to define baseline behavior (e.g., a default RS policy) against which to measure the ``loss'' incurred by an individual.
\end{itemize}
Naturally, there are many other factors that can be incorporated into the design of an SCF.

The use of an SCF  presents numerous challenges for specification, elicitation, assessment, measurement, and optimization as discussed in the preceding sections. For instance, most RSs rarely take steps to assess true user, creator or vendor utility, instead relying on proxy measurements (e.g., return engagement). From an optimization perspective, the fact that agent utility should generally be measured over extended horizons itself presents challenges, since RS rarely optimize policies non-myopically.

Another challenge pertains to interpersonal comparison of utilities \cite{harsanyi_cardinal:1955}. Abstractly, an SCF simply maps preferences into RS action/policy choices. In practice, however, one generally assumes \emph{quantitative agent utilities} overs outcomes, defines a social welfare function that \emph{scores} outcomes as a function of these utilities, and selects the choice that is score (or welfare) maximizing.
As such, the tradeoffs in utility of one agent (or group) against that of another embodied in the SWF must assume that these utilities (as estimated or elicited by the RS) are comparable. 

The need for quantitative utility functions is especially acute in RSs, where learning methods are commonly used to assess quantitative user affinities, and expectations must be taken over uncertain outcomes. Moreover, boiling things down into individual utilities and assuming comparability provides significant traction w.r.t.\ optimization. But it remains to be seen what one loses in terms of generality. We note that in much of mechanism design, the use of monetary transfers and the assumption of quasi-linear utility obviates this concern (e.g., this assumption underlies most of auction theory). As discussed above, however, mechanism design within RSs will often fall within the realm of \emph{mechanism design without money} \cite{SchummerVohra:facilityLocSurvey,procacciaTennen:acmec09}.

Questions about disparate treatment of actors are often treated within the domain of ML fairness. Classical notions of fairness are concerned with equalising the performance of a machine learning model amongst groups or individuals by correcting for various data or modelling biases. In RSs, these concerns are reflected both on the user and creator side \cite{ekstrand_fair_recsys:handbook22,li_etal:recs_fairness}, where the goal is generally to achieve more equitable model quality of user preferences/item features across various protected groups. Another form of fairness is that of fairness of exposure, which deals with the fair distribution of attention to creators, usually based on their estimated quality, by manipulating the ranking of recommended items. All of these approaches can be seen as implicit attempts to promote certain social outcomes over others, but the relationship between adjusting models and actual utility of outcomes can sometimes be hard to quantify. Therefore, SCFs can be seen as a more powerful language to address these issues at the level of outcomes, rather than interventions on models.  



\section{Social Objectives}
\label{sec:socialSCF}

Some elements of the SCF may not be readily definable in terms of the utilities or preferences of the individual actors. This may be the case, for instance, when the SCF involves properties of the \emph{joint} outcome over, say, users, but where each user is concerned only about their ``local'' part of that outcome. One such phenomenon with this property, commonly discussed in relation to RSs, is that of \emph{filter bubbles}, and the potential for induced social or intellectual fragmentation across the user population \cite{pariser_filter_bubble_book:2011,aridor_filterbubble:recsys20}. This may emerge when the preferences of individual users do not explicitly value diversity w.r.t.\ some types of content.\footnote{This preference attitude may be due to general indifference, or may be an active social preference, e.g., as in the case of a user's perception of being an influential member of a community, and the associated sense of belonging/identity \cite{klein_polarized_book:2020}.} Thus, different user subpopluations, when engaging with an RS that can effectively maximize user utility, may end up consuming different collections of content.

Whether such filter bubbles are problematic, of course, depends on the nature of the content in question. A ``filter bubble'' in which one groups of users largely consumes music from the sub-genre electro-house, while another listens to indie folk will should not be considered problematic. But if the content, say, involves diverse perspectives on social, scientific, civic or political issues of significant social import, the fact that large groups of users have no exposure to each other's perspectives may be of greater concern. This might be best viewed as a societal value, where these consumption patterns have a negative societal externality, e.g., preventing meaningful civic discourse. Penalizing such patterns (or assigning utility to at least some degree of overlap in consumption) to minimize these externalities within the SCF (i.e., as part of the RS's long-term objective) provides one means for encoding such  considerations. Other phenomena of this sort include polarization and radicalization \cite{Ribeiro_etal:FAT20, musco_etal:www18}, personal diversity of consumption, ``echo chambers,'' and the like.\footnote{Of course, if one believes that these are harmful externalities for the RS's users themselves, in principle, one could encode these considerations in the utility functions of the users; but doing so would require a deeply sophisticated model of long-term outcome dynamics. Hence, simpler proxies involving overlap and/or diversity of content consumption should prove more actionable in practical recosystems.}

Finally, it is worth drawing attention to the interaction between social objectives and fairness. For example, rich-get-richer feedback loops---which can easily lead to heavy-tailed distributions of content/creator popularity, influence or income---are typically seen as undesirable from the point of view of fairness. Yet, economies of scale often have the property that high production-value content can be more easily produced by a single creator with access to greater resources or budget than by ten creators, each with one-tenth of the budget, under conditions of competition. Moreover, having a few items of content with out-sized popularity may be inherently socially beneficial as ``cornerstone content'' for building genres and communities. It is easy to construct hypothetical scenarios where fairness and other societal values are at odds. This too suggests that SCFs should be taken seriously as a means for addressing fundamental tradeoffs in recosystems.



\section{The Elephant in the Room}
\label{sec:SCFelephant}

One of the most fundamental questions in the use of mechanism design to design and optimize recosystems is the selection of the SCF. While akin to the selection of the objective in any optimization problem, or loss function in any machine learning problem---especially those of a multi-objective or multi-task nature---deciding on an SCF carries with it additional ``baggage,'' considerations that (usually) play no role in ML or traditional multi-objective problems.

One distinction is that tradeoffs are being made over the utility generated for different individuals. This stands in contrast to single-party multi-objective optimization, where tradeoffs across objectives lie in the hands, and only impact, a single decision making entity. The self-interested agents impacted by the RS policy may not agree with the tradeoffs being made (see more below). This ultimately raises important questions regarding who should help shape the nature of the SCF, and who should make the final determination? Such considerations are rendered more complex still if one is to account for social objectives of the form discussed in Section~\ref{sec:socialSCF}.

A second factor are the seminal impossibility results which make the considerations above even more difficult.  For instance, Arrow's \shortcite{arrow50} impossibility theorem makes clear that appeal to some ``universal'' SCF that embodies incontrovertible principles is pointless.\footnote{Arrow's result states that, except in some simple settings, no SCF (i.e., preference aggregation scheme) satisfies all of a set of (what are generally viewed as) reasonable conditions. This mathematical impossibility does not mean aggregation doesn't work; it simply means that, for any (non-dictatorial) SCF, there exists \emph{some} configuration of preferences where the SCF will do something considered to be unreasonable.} Even seemingly uncontroversial conditions like Pareto optimality can sometimes be called into question on philosophical grounds \cite{sen_impossibility:jpl1970}.

Apart from deciding how to aggregate preferences to make collective choices, the question of accurately eliciting or estimating the preferences of the participants in the RS ecosystem is yet a third distinguishing factor. Indeed, the self-interested nature of the recosystem participants renders this especially challenging. The classic Gibbard-Satterthwaite \cite{gibbard:1973,satterthwaite:1975} impossibility result states that no mechanism can incentivize all participants to reveal their preferences truthfully (except, again, in relatively simple settings). This result may be of marginally less practical import than Arrow's, since strategic misrepresentation or misreporting of preferences can be both computationally and informationally difficult, and can often have a relatively limited impact on outcome quality \cite{LTPB:uai12}; but the possibility of strategic misreporting should still play a role in the selection of a suitable SCF. 

As a result, determining a suitable SCF may be one of the most daunting tasks in the application of mechanism design principles to the design of recosystems. For \emph{any} given SCF, simply predicting its ultimate impact on the \emph{realized} outcomes of an RS policy, and the \emph{realized} expected utility generated for ecosystem players (and their tradeoffs), is tremendously complex. The development of analytical, simulation, visualization and scenario-analysis tools seems vital to help in exploring the space of SCFs. This should also play a critical role in the transparency of RS designs and their consequences for individuals and societal groups.

\part{Concluding Remarks}
\label{part:conclude}

We have defined an ambitious research program, consisting of a set of challenging---but potentially immensely impactful---research problems in the arena of recommender ecosystems. Given the pervasive nature of recommender systems, and their ever-increasing scope and influence on our daily lives, success in addressing these research challenges will have broad implications for scientists studying RSs, practitioners who deploy RSs, and on societal dynamics and values.

\bibliography{long,standard,newrefs}

\end{document}